# Novel Span Measure, Spanning Sets and Applications

Nidhika Yadav[a]

**Abstract:** Rough Set based Spanning Sets were recently proposed to deal with uncertainties arising in the problem in domain of natural language processing problems. This paper presents a novel span measure using upper approximations. The key contribution of this paper is to propose another uncertainty measure of span and spanning sets. Firstly, this paper proposes a new definition of computing span which use upper approximation instead of boundary regions. This is useful in situations where computing upper approximations are much more convenient that computing boundary region. Secondly, properties of novel span and relation with earlier span measure are discussed. Thirdly, the paper presents application areas where the proposed span measure can be utilized.

## 1. Introduction

Rough Sets [16, 17] have been used for dealing with decision making in uncertainty. It has wide range of applications not limiting to feature selection [2, 11,12, 19], classification [13], finance [18], clustering [15], data mining [8, 14], information retrieval [4], text summarization [20, 21]. The primary use of Rough Sets have been in problems that increase the accuracy of the supervised systems. Recently Yadav et al [21[ proposed its use in unsupervised learning as well. This opens doors to lot of unsupervised learning problems not limiting to NLP only.

Yadav et al (2019) [21] proposed the definition of span and spanning set for an information system. Typically, such an information system is represented via an information table. Rough Set based lower and boundary region were used to compute the span of any subset of a Universe. The span depends on a trade-off between possibly covered concepts and surely covered concepts which are computed with Rough Set based approximations. Span defines the spanning or covering capabilities of a subset of universe. It gives a measure in percentage how much of the objects of universe this subset covers given the knowledge granules. The definition of span is given as follows (Yadav et al., 2019) [21].

**Definition 1**. Given a universe $U$ and an Information System $(U, R)$. The span of a subset $X$ of $U$ for a subset $P$ of $R$ is defined as:

$$\delta_{X,P} = \left(w_1 * \frac{|PX|}{|U|} + w_2 * \frac{|BN_P(X)|}{|U|}\right), \text{ where } w_1, w_2 \in [0,1], w_1+w_2=1.$$

The subset of universal set that maximizes the span is referred to as spanning set. The spanning set is obtained by solving unconstrained optimization problem:

$$maximize\ (\delta_{P,Z}) \qquad\qquad\qquad\qquad\qquad\qquad\qquad\qquad\qquad\qquad\qquad\ldots\ldots\ldots\ldots\ldots\ldots(1)$$

[a] email address: nidhikayadav678@outlook.com

The spanning set of size m is obtained by solving the following optimization problem:

$$maximize\ (\delta_{P,Z})$$

$$subject\ to,\ length(Z) \leq m$$

The reason for the name, spanning set, is very evident, this set covers most of the information granules present in the partition of the universe by the equivalence relation. Further, the concept is applied on real world problems in the area of natural language processing by authors [21]. The application was well elaborated by the authors [21] and mathematical concepts were illustrated. However, this paper argues that another parallel definition can be used with in certain cases. In the next section proposed new definition of span is introduced, its utility, properties and applications are described.

The new definition of span utilizes upper approximation degree rather than boundary region. This kind measure when optimized either via constrained or unconstrained optimization techniques can find solutions to problems that lay emphasis on upper approximations and more on lower approximations. These set of problems have altogether a different application area. Two applications have been discussed in this paper. Firstly, application in Fuzzy Rough Sets, the upper approximation and lower approximations are readily available in such applications, and hence a perfect candidate to the proposed approach. Secondly, it is mathematically proved that the optimized solution so formed leads to a higher accuracy, hence can be applied to problem of Rough Set based Feature Selection.

Firstly, we introduce the concept of Fuzzy Rough Sets [3, 11, 12, 13]. Let us consider a universe U. Let R be fuzzy binary relation on U, satisfying the following, (a) $R(x,x) = 1$, reflexive relation (b) $R(x,y) = R(y,x)$, symmetric property and (c) $R(x,y) \geq min_z \{R(x,z), R(z,y)\}$, the transitive property. The three prominent definitions of Fuzzy Rough Sets are as follows [3, 11, 12, 13]:

I. Given a Fuzzy Set µ and Fuzzy Equivalence Relation R, the Fuzzy lower and Fuzzy upper approximations are defined as:
$$\underline{R}F(x) = \min\{\max\{1 - R(x,y), \mu(y)\}, y \in U\}$$
$$\overline{R}F(x) = \max\{\min\{R(x,y), \mu(y)\}, y \in U\}$$

II. The upper and lower approximations for a Fuzzy Set F are given as follows:
$$\mu_{\underline{R(X)}}(F) = \min\{\max\{1 - \mu_F(y), \mu_X(y)\}: y\ in\ U\}$$
$$\mu_{\overline{R(X)}}(F) = \max\{\min\{\mu_F(y), \mu_X(y)\}: y\ in\ U\}$$

III. An approach to compute the memberships as follows:
$$\mu_{\underline{R(X)}}(x) = sup_{F \in U/P} \min\{\mu_F(x), \min\{\max\{1 - \mu_F(y), \mu_X(y)\}: y\ in\ U\}\}$$
$$\mu_{\overline{R(X)}}(x) = sup_{F \in U/P} \min\{\mu_F(x), \max\{\min\{\mu_F(y), \mu_X(y)\}: y\ in\ U\}\}$$

The applications of Fuzzy Rough Sets are diverse and it fits in many real time problems especially involving real valued data. As the key drawback of Pawlak Rough Sets [16, 17] is that is require categorical data which is not need in Fuzzy Rough Sets. The following sections formally define the new span measure for Rough Sets.

The motivation of this paper is to propose a novel Rough Set based uncertainty measure that can be used in supervised as well as unsupervised environments. And this measure overcomes the drawback of previously defined span measure in especially problems where boundary region is not easily computed. The paper is organized as follows. Section 2 described the proposed terminologies and proposed uncertainty measure. Properties of proposed novel uncertainty measure are discussed in Section 3. Section 4 illustrates a real life example of the proposed measure in computing span measure of Fuzzy Rough Sets and also define an equivalent definition which can be used in place of the proposed definition. Both use the upper approximations. Section 5 concludes the paper with future works.

## 2. Proposed Terminologies

In this paper another definition to span is proposed. The motivation behind this approach is ease of use, necessity and applications. There are times when boundary region may not be computable as readily as the upper approximation and at time not at all computable. There are applications where computing boundary region require computing of upper approximation first. At times efficiency of computational time is quite important and boundary region computation may require extra overhead in computations. There are times when upper approximation is more suitable to the problem solving. And in some problems upper approximation is computed readily, especially in variants of Rough Sets, as can be seen in later part of the article. Hence, it is proposed in this paper to also define covering or spanning capabilities of a subset of a universe using upper approximation instead of just the boundary region. Here, is the proposed definition of span based on lower approximation and upper approximations. The proposed definition is as follows.

**Definition 2 (Span).** Given a universe $U$ and an Information System $(U, R)$. The span of a subset $X$ of $U$ w.r.t subset $P$ of R is defined as:

$$\delta'_{X,P} = \left(w_1 * \frac{|\underline{R}X|}{|U|} + w_2 * \frac{|\overline{R}X|}{|U|}\right), \text{ where } w_1, w_2 \in [0,1], w_1+w_2=1.$$

It is to be emphasised here that upper approximation is used here instead of boundary region. In the following propositions the relation between the span measure of a subset of universe from Definition 1 and Definition 2 is established.

The subset of universal set that maximizes the span is referred to as spanning set. The spanning set is obtained by solving unconstrained optimization problem:

$$maximize\ (\delta'_{P,Z}) \qquad \ldots\ldots\ldots\ldots(2)$$

While a spanning set with a constraint of size can be obtained by making (2) a constrained optimization problem, where limit of length is set.

**Proposition 1.** The following relationship holds between the two definitions of span.

i. $\delta'_{X,P} = w_2 \frac{|\overline{P}X|}{|U|} + \delta_{X,P}$,
ii. $\delta'_{X,P} \geq \delta_{X,P}$

Hence the spanning set obtained by maximizing the new definition of span lays more emphasis on lower approximations.

**Proof.** Let $X$ be a subset of $U$ and span $\delta'_{X,P}$ be defined as:

$$\delta'_{X,P} = \left(w_1 * \frac{|\underline{P}X|}{|U|} + w_2 * \frac{|\overline{P}X|}{|U|}\right), \text{ where } w_1, w_2 \in [0,1], w_1+w_2=1$$

And let the original definition of span produce span $\delta_{X,P}$, where $\delta_{X,P} = \left(w_1 * \frac{|\underline{P}X|}{|U|} + w_2 * \frac{|BN_P(X)|}{|U|}\right)$, where $w_1, w_2 \in [0,1]$, $w_1+w_2=1$.

Now, $\delta'_{X,P} = \left(w_1 * \frac{|\underline{P}X|}{|U|} + w_2 * \frac{|\overline{P}X|}{|U|}\right) = \left(w_1 * \frac{|\underline{P}X|}{|U|} + w_2 * \frac{|\underline{P}X|+|BN_P(X)|}{|U|}\right)$

$= \left((w_1 + w_2) * \frac{|\underline{R}X|}{|U|} + w_2 * \frac{|BN_P(X)|}{|U|}\right)$

$= w_2 \frac{|\underline{P}X|}{|U|} + w_1 * \frac{|\underline{P}X|}{|U|} + w_2 * \frac{|BN_P(X)|}{|U|}$

$= w_2 \frac{|\underline{P}X|}{|U|} + \delta_X$, hence proved, (i) follows form this relation.

Hence the new span lays more emphasis on lower approximation and at the same time does not ignore the boundary region. The applications of this concept are described in Section 5. The result is evident from the definition new measure for span, that has upper approximation which inherently includes the lower approximation. An analogous of Definition 2 can be made for a complete subset R of the Information System (U, R). This is given as follows.

**Definition 3.** Given an information system (U, R). Let X be a subset of U. The span of X is the weighted average,

$\delta'_X = \left(w1 \frac{|\underline{R}X|}{|U|} + w2 \frac{|\overline{R}X|}{|U|}\right)$, w1, w2 $\in$ [0,1], w1+w2 =1.

Similar result as in Proposition 1 hold for $\delta'_{P,X}$ as well. Definition 2 and definition 3 of span is useful in situations and problems were computing upper approximation is much more convenient than computing boundary regions. For example, as we shall illustrate the case below for Fuzzy Rough Sets where lower and upper approximations are computed, rather than boundary region. Spanning set is a subset Y of $U$ which maximizes the span. And similar definition of spanning set follows for the new measure of span, which is proposed above.

# 3. Some Mathematical Properties of Proposed Span Measure

The new definition of span satisfy different properties that span proposed by Yadav et al.(2019). This section provides a glimpse of some mathematical results on span results based on new definition proposed in this paper. Let Q ⊆ P ⊆ A. Then, it can be proved that $\underline{P}X \supseteq \underline{Q}X$ and $\overline{P}X \subseteq \overline{Q}X$. Also, the following propositions hold for the new definition of span. These can be proved with little efforts, only important results proofs are provided in the paper.

**Proposition 2.** Suppose (U, A) be an information system. For a subset X ⊆ U and attribute subsets P, Q⊆A, such that Q⊆P, then $\delta'_{P,X} \leq \delta'_{Q,X} + \theta$, where $\theta = w1 * |BN_Q(X)| / |U|$. Also, when $\lim_{w1 \to 0} \theta = 0$, $\delta'_{P,X} \leq \delta'_{Q,X}$.

**Proposition 3.** Suppose (U, A) be an information system. For an object subset X⊆U and attribute subsets P, Q ⊆ A, such that Q ⊆ P, and $\underline{P}X = \underline{Q}X$, then $\delta_{P,X} \leq \delta_{Q,X}$.

**Proposition 4.** If X is a span of (U, A) and Y arbitrary set of U such that |X| = |Y|, then for any P, $\delta_{P,Y} \leq \delta_{P,X}$.

**Proposition 5.** Let Z be a span of (U, A) and let Y be any other subset of U. Then, the following inequality holds,

$$|\overline{P}Y| - |\underline{P}Z| \leq |\overline{P}Z| - |\underline{P}Y|$$

**Proposition 6.** Let w1, w2 be such that w1+ w2 = 1 and w1 ≥ w2. Let another set of weights be u1 and u2, such that u1 + u2 = 1, and u1 ≥ u2. Then for fixed P and X. Let, $\delta'_{P,X} = (u1 \frac{|\underline{P}X|}{|U|} + u2 \frac{|\overline{P}X|}{|U|})$ and $\delta_{P,X} = (w1 \frac{|\underline{P}X|}{|U|} + w2 \frac{|\overline{P}X|}{|U|})$. Then, when u1 > w1 then $\delta'_{P,X} \leq \delta_{P,X}$ and when u1 < w1 then $\delta'_{P,X} \geq \delta_{P,X}$.

The following proposition is important since it proves that the maxima of the problem (1) with first span measure is never better than the optimal value of problem (2) with proposed span measure.

**Proposition 7.** Given an Information System $(U, R)$ and let Z be the spanning set obtained by using span $\delta$ and let $Z'$ be spanning set obtained by span $\delta'$. Then, $\delta_{Z,P} \leq \delta'_{Z',P}$.

**Proof.** Let Z is spanning set w.r.t. span measure $\delta$ and let $Z'$ be spanning set w.r.t span measure $\delta'$

Then, $\delta'_{Z,P} = w_2 \frac{|\overline{P}Z|}{|U|} + \delta_{Z,P}$

⇨ $\delta'_{Z,P} \geq \delta_{Z,P}$, for fixed weights w1 and w2.  ……(3)

$\delta'_{Z',P} = w_2 \frac{|\overline{P}Z'|}{|U|} + \delta_{Z',P}$

⇨ $\delta'_{Z',P} \geq \delta_{Z',P}$, for fixed weights w1 and w2.  ……(4)

Z and $Z'$ are subsets of universe U. Hence by definition od spanning sets. The following inequalities hold, as spanning set is the subset of universe which maximizes the corresponding measures.

$\delta'_{Z,P} \leq \delta'_{Z',P}$, $Z'$ spanning set w.r.t. measure $\delta'$ and Z a subset of universe.  ……(5)
$\delta_{Z',P} \leq \delta_{Z,P}$, Z spanning set w.r.t. measure and $Z'$ a subset of universe.  ……(6)

From (3), (4), (5) and (6) it follows that

$\delta'_{Z,P} \geq \delta_{Z,P} \geq \delta_{Z',P}$  from (3) and (6)  ……(7)
$\delta'_{Z,P} \leq \delta'_{Z',P} \geq \delta_{Z',P}$  from (4) and (5)  ……(8)
$\delta_{Z,P} \leq \delta'_{Z,P} \leq \delta'_{Z',P}$

From (7) and (8) it follows that, $\delta_{Z,P} \leq \delta'_{Z',P}$

It follows the maximum attained by the exact spanning set of Definition 1 of span $\delta$ is always less than span measure of spanning set obtained by Definition 2 of span $\delta'$.

**Proposition 8.** Given an Information System $(U, R)$. Let $X \subseteq U$, $P \subseteq R$ and $Z'$ be a spanning set satisfying $|\overline{P}Z'| \leq |\overline{P}X|$. Then the following holds:

(i) The accuracy given by $\alpha_P(X) = \frac{|\underline{P}X|}{|\overline{P}X|}$, satisfies $\alpha_P(X) \leq \alpha_P(Z')$

(ii) The roughness given by $\rho_P(X) = 1 - \alpha_P(X)$, satisfies $\rho_P(X) \leq \rho_P(Z)$.

**Proof.** Let $Z'$ be a spanning set w.r.t. span $\delta'$ and X be a subset of U. Then

$$\delta'_{X,P} = \left( w_1 * \frac{|\underline{P}X|}{|U|} + w_2 * \frac{|\overline{P}X|}{|U|} \right)$$

$$\delta'_{Z',P} = \left( w_1 * \frac{|\underline{P}Z'|}{|U|} + w_2 * \frac{|\overline{P}Z'|}{|U|} \right)$$

Since $Z'$ is spanning set, the following holds,

$\delta'_{X,P} \leq \delta'_{Z',P}$,

$\left( w_1 * \frac{|\underline{P}X|}{|U|} + w_2 * \frac{|\overline{P}X|}{|U|} \right) \leq \left( w1 * \frac{|\underline{P}Z'|}{|U|} + w2 * \frac{|\overline{P}Z'|}{|U|} \right)$

$\Rightarrow \frac{|\overline{P}X|}{|U|} \left( w_1 * \frac{|\underline{P}X|}{|\overline{P}X|} + w_2 * \frac{1}{|U|} \right) \leq \frac{|\overline{P}Z'|}{|U|} \left( w1 \frac{|\underline{P}Z'|}{|\overline{P}Z'|} + w2 * \frac{1}{|U|} \right)$

$\Rightarrow \left( w_1 * \frac{|\underline{P}X|}{|\overline{P}X|} + w_2 * \frac{1}{|U|} \right) \leq \frac{|\overline{P}Z'|}{|\overline{P}X|} \left( w1 \frac{|\underline{P}Z'|}{|\overline{P}Z'|} + w2 * \frac{1}{|U|} \right)$

$\Rightarrow \left( w_1 * \frac{|\underline{P}X|}{|\overline{P}X|} + w_2 * \frac{1}{|U|} \right) \leq \left( w1 \frac{|\underline{P}Z'|}{|\overline{P}Z'|} + w2 * \frac{1}{|U|} \right)$, by the given assumption

$\Rightarrow \left( w_1 * \frac{|\underline{P}X|}{|\overline{P}X|} \right) \leq w1 \frac{|\underline{P}Z'|}{|\overline{P}Z'|}$

$\Rightarrow \left( \frac{|\underline{P}X|}{|\overline{P}X|} \right) \leq w \frac{|\underline{P}Z'|}{|\overline{P}Z'|}$

$\Rightarrow \alpha_P(X) \leq \alpha_P(Z')$

$\Rightarrow \rho_P(Z') \leq \rho_P(X)$

Proposition 8 is an important Proposition, which means that the accuracy of the system increases with the increase in span provided the boundary regions is constrained, while the roughness decreases. This proves that this measure of span can be used for problems where classification accuracy is needed to be higher as the algorithm progresses. One such example is feature selection problem. Hence, in applications section we propose use of new span measure for feature selection.

From the results above it follows that span measure by a smaller attribute set is more than the span measure by a larger attribute set. Hence, the complete span for $\delta'$ is proposed to be defined as follows:

$$\delta'^{Complete}_{P,X} = \sum_{ai \in P} \delta'_{ai,X}$$

The complete span misses out the effect of full attribute set. This motivated us to define the complete span as follows.

**Definition 3.** The span is subset X of U based on complete set of discerning attributes is given by:

$$\delta'^{Complete}_{P,X} = \sum_{ai \in P} \delta'_{ai,X} + \delta_{P,X}$$

The difference between the complete span of Yadav et al (2019) and the proposed complete span is two-fold. Firstly, it lays emphasis on the approximations computed by the complete attribute subsets and by the individual attributes as well. Secondly, it uses the new measure for computing span. The following are the three ways to compute spanning sets of either kind for the new span measure $\delta'$.

a) Rough Set based Spanning Set,
$$maximize \quad \delta'_{P,X}$$

b) Complete Rough Set based Spanning Set,
$$maximize \quad \sum_{ai \in P} \delta'_{ai,X}$$

c) Hybrid Spanning Set,
$$maximize \quad \sum_{ai \in P} \delta'_{ai,X} + \delta'_{P,X}$$

## 4. Applications of Novel Span Measure

In these section two applications of the proposed span using upper approximation have been illustrated.

### 4.1. Application in Fuzzy Rough Sets

Below, we provide the definition of span and spanning sets for Fuzzy Rough Sets, a variant of Rough Sets. Fuzzy Rough Set based span measure is defined as follows:

$$\delta'_{X,P} = \left(w_1 * \frac{|\underline{P}X|}{|U|} + w_2 * \frac{|\overline{P}X|}{|U|}\right)$$

Since, we are dealing with Fuzzy Rough Sets, hence each of the lower and upper approximation is a Fuzzy Set and hence the definition can also be written as:

**Definition 4.** Given a Fuzzy Rough Set the span measure defined as:

$$\delta'_{P,X} = (w1 \frac{\left|\sum \mu_{\underline{R(X)}}(x)\right|}{|U|} + w2 \frac{\left|\sum \mu_{\overline{R(X)}}(x)\right|}{|U|})$$

It can be proved that these both are the same definition, by the very definition of cardinality of a Fuzzy Set [5].

**Example 1.** Let us illustrate with an example from Natural Language Processing. Consider the following sentences.
  i. Sentence 1, S1 = 'A group of boys in a yard is playing and a man is standing in the background'
  ii. Sentence 2, S2 = 'A group of kids is playing in a yard and an old man is standing in the background'

Consider a universe U as the words in the essential words in the domain. Here U, for example, can be taken as, U = ['kids', 'group', 'standing', 'boys', 'background', 'old', 'man', 'yard', 'playing']

The following definition is used for the Fuzzy lower and Fuzzy upper approximations computations.
$$\underline{RF}(x) = \min\{\max\{1 - R(x,y), \mu(y)\}, y \in U\}$$
$$\overline{RF}(x) = \max\{\min\{R(x,y), \mu(y)\}, y \in U\}$$

Here, R can be computed using traditional gaussian distance between word vectors or a similarity measure. Consider for example using a WordNet based similarity measure. Then using above formulas, the lower and upper approximations for Sentence 1 and Sentence 2 are computed as follows:

Lower approximation of Fuzzy Set of Sentence 1 is [0.6667, 0.5556, 0.5455, 0.6667, 0.5455, 0.5455, 0.6667, 0.6923, 0.7647]

Upper approximation of Fuzzy Set of Sentence 1 is [0.6667, 1, 1, 1, 1, 0.5455, 1, 1, 1]

Lower approximation of Fuzzy Set of Sentence 2 is [0.6667, 0.7778, 0.8334, 0.6667, 0.8182, 0.8334, 0.6667, 0.8462, 0.8823]

Upper approximation of Fuzzy Set of Sentence 2 is [1, 1, 1, 0.6667, 1, 1, 1, 1, 1]

The span measure for Sentence 1 and Sentence 2 using the proposed definition in paper is given in Table 1.

| Weights | Span Sentence1 | Span Sentence2 |
|---|---|---|
| w1=0.2, w2=0.8 | 0.8554 | 0.9257 |
| w1=0.1, w2=0.9 | 0.8839 | 0.9443 |
| w1=0.8, w2=0.2 | 0.8554 | 0.9257 |
| w1=0.9, w2=0.1 | 0.6561 | 0.7954 |
| w1=0.5, w2=0.5 | 0.7700 | 0.8698 |
| w1=0.3, w2=0.7 | 0.8270 | 0.9071 |
| w1=0.7, w2=0.3 | 0.7130 | 0.8326 |

Table 1. Span measure for Sentence 1 and Sentence 2 using the proposed definition using various combination of weights

From this example and computations, it is clear that given this domain, Sentence 2 gives a consistently higher score of novel span measure than Sentence 1, whichever values of weights be chosen. This information can be used in the problem of ***word sense disambiguation***. Here, a domain and hence a universe of terms are fixed, the span measure can be computed to fit in the lemmas of words given in the dictionary. Whereas the universe is defined as the words in the sentence surrounding it. The lemma of which ever word has a higher span measure can be considered as its actual sense. The experiments shall be performed as a part of future work. This paper is for proposal and illustration.

### 4.2. Application in Feature Selection.

The intuition behind this application lies in the fact that more emphasis is given to lower approximation and hence on accuracy. This can be has been proved mathematically above in paper that the accuracy increases with the increase in new span measure, given a constraint on boundary region. Rough Set has been popularly used for task of feature selection [2. 11. 12, 19]. The application typically utilize only the lower approximations, reducts and positive region. However, experiments can be conducted to evaluate the application of proposed technique to feature selection problem, since the Proposition 8 proves that accuracy increases with the increase in proposed span measure value. And hence can guide the algorithm in a better state space search, given some amount of emphasis is given to boundary region as well. This is left as future work.

## 5. Conclusion and Future Works

This paper proposes another uncertainty measure called span which lays more emphasis on lower approximations, not through weights but through inclusion of upper approximation in computing the novel span measure instead of boundary region. The earlier proposed span measure had its own advantageous. While, the use and advantages of the proposed span measure has been well elaborated in the paper. Further, as future work, the proposed applications have to be experimentally verified for use over standard datasets in various problems such as word sense disambiguation and feature selection, which have been proposed in this paper. Further, the optimization of weights is a problem dependent on application.